\documentclass[11pt,a4paper]{article}
\usepackage[hyperref]{emnlp2020}
\usepackage{times}
\usepackage{latexsym}
\usepackage{amsmath,amssymb}
\usepackage{enumitem}
\usepackage{tikz}
\usepackage[linguistics]{forest}
\usetikzlibrary{arrows}

\usepackage{multirow}

\usepackage{microtype}

\aclfinalcopy 

\title{Examining the rhetorical capacities of neural language models}

\author{Zining Zhu$^{1,2}$, Chuer Pan$^1$, Mohamed Abdalla$^{1,2}$, Frank Rudzicz$^{1,2,3,4}$ \\
  1: University of Toronto; 2: Vector Institute \\
  3: Li Ka Shing Knowledge Institute, St Michael's Hospital \\
  4: Surgical Safety Technologies\\
  \texttt{zining@cs.toronto.edu, chuer.pan@mail.utoronto.ca}\\
  \texttt{\{msa, frank\}@cs.toronto.edu} \\
  }

\date{}

\begin{document}
\maketitle
\begin{abstract}
Recently, neural language models (LMs) have demonstrated impressive abilities in generating high-quality discourse. While many recent papers have analyzed the syntactic aspects encoded in LMs, to date, there has been no analysis of the inter-sentential, rhetorical knowledge. 
In this paper, we propose a method that quantitatively evaluates the rhetorical  capacities of neural LMs. We examine the capacities of neural LMs understanding the rhetoric of discourse by evaluating their abilities to encode a set of linguistic features derived from Rhetorical Structure Theory (RST). Our experiments show that BERT-based LMs outperform other Transformer LMs, revealing the richer discourse knowledge in their intermediate layer representations. In addition, GPT-2 and XLNet apparently encode less rhetorical knowledge, and we suggest an explanation drawing from linguistic philosophy. Our method presents an avenue towards quantifying the rhetorical capacities of neural LMs.
\end{abstract}

\section{Introduction}
In recent years, neural LMs (especially contextualized LMs) have shown profound abilities to generate texts that could be almost indistinguishable from human writings \citep{radford2019language}. Neural LMs could be used to generate concise summaries \citep{song2019mass}, coherent stories \citep{see-etal-2019-massively}, and complete documents given prompts \citep{keskarCTRL2019}. It is natural to question their source and extent of rhetorical knowledge: What makes neural LMs articulate, and how? 
While some recent works query the linguistic knowledge \citep{hewitt-manning-2019-structural-probe,Liu2019,Chen2019,Belinkov2017}, this open question remain unanswered. We hypothesize that contextualized neural LMs encode rhetorical knowledge in their intermediate representations, and would like to quantify the extent they encode rhetorical knowledge.

To verify our hypothesis, we hand-craft a set of 24 rhetorical features including those used to examine rhetorical capacities of students \citep{Mohsen2019,liu2016write-to-learn,Zhang2013ETS-report,Powers2001}, and evaluate how well neural LMs encode these rhetorical features in the representations while encoding texts.

Recent work has started to evaluate encoded features from hidden representations. Among them, probing \citep{Alain2017,adi2017fine} has been a popular choice. Previous work probed morphological \citep{Belinkov2017,bisazza-tump-2018-lazy}, agreement \citep{giulianelli-etal-2018-hood}, and syntactic features \citep{hewitt-manning-2019-structural-probe,hewitt-liang-2019-designing}. Probing involves optimizing a simple projection model from representations to features. The loss of this optimization measures the difficulty to decode features from the representations. 

In this work, we use a probe containing self attention mechanism. We first project the variable-length embeddings to a fixed-length latent representation per document. Then, we apply a simple diagnostic classifier to detect rhetorical features from this latent representation. This design of probe reduces the total number of parameters, and enable us to better understand each model's ability to encode rhetorical knowledge.
We find that:
\begin{itemize}[noitemsep]
    \item The BERT-based LMs encode more rhetorical features, and in a more stable manner, than other models.
    \item The semantics of non-contextualized embeddings also pertain to some rhetorical features, but less than most layers of contextualized language models.
\end{itemize}

These observations allow us to investigate the mechanisms of neural LMs to better understand the degree to which they encode linguistic knowledge. We demonstrate how discourse-level features can be queried and analyzed from neural LMs. All of our code and parsed tree data will be available at github.

\section{Structural analysis of discourse}
Various frameworks exist for ``good discourse'' \citep{Lawrence2019ArgumentMiningSurvey,irish2009engineering,Toulmin1958}, but most of them are inaccessible to quantitative analysis. In this work, we use Rhetorical Structure Theory \citep{Mann1988,Mann1989RST} since it represents the structures of discourse using trees, allowing straightforward quantitative analysis.
There are two components in an RST parse-tree:
\begin{itemize}[noitemsep]
    \item Each leaf node represents an elementary discourse unit (EDU). The role of an EDU in an article is similar to that of a word in a sentence.
    \item Each non-leaf node denotes a relation involving its two children. Often, one of the children is more dependent on the other, and less essential to the writer's purpose. This child is referred to as ``satellite'', while the more central child is the ``nucleus''.
\end{itemize}

\begin{figure}[h]
\centering
\begin{forest}
[NS-Contrast,rectangle,draw
    [SN-Attribution,rectangle,draw,name=AB
        [I didn't\\ know,name=A]
        [this is \\from C,name=B]]
    [but it is\\very good!,name=C]
]
\end{forest}
\caption{A portion of an RST tree, selected from IMDB \citep{Maas2011IMDB} \texttt{train/pos/1\_7.txt}, and parsed with \citet{feng-hirst-2014-linear}. Nodes with rectangle borders are discourse relations, and those without borders are individual EDUs. The ``N'' and ``S'' prefix for discourse relations stand for ``nucleus'' and ``satellite'' respectively.
\label{fig:rst-tree}}
\end{figure}
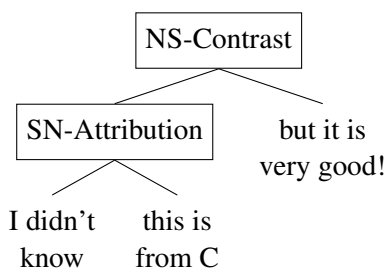

Tree representations are clear, easy to understand, and allow us to compute features to numerically depict the rhetorical aspects of documents.

\subsection{Rhetorical features}
\label{subsec:rhetorical-features}
Previous work used RST features to analyze the quality of discourse, to assess writing abilities \citep{wang2019RST-ETS,Zhang2013ETS-report}, examine linguistic coherence \citep{feng-hirst-2014-coherence,Abdalla2017}, and to analyze arguments \citep{chakrabarty-etal-2019-ampersand}. In this project, we extract similar RST features in the following three categories:

\paragraph{Discourse relation occurrences (\texttt{Sig})} We include the number of relations detected in each document. There are 18 relations in this category\footnote{The 18 relations are: Attribution, Background, Cause, Comparison, Condition, Contrast, Elaboration, Enablement, Evaluation, Explanation, Joint, Manner-Means, Topic-Comment, Summary, Temporal, Topic-Change, Textual-organization, and Same-unit.}. 
Unfortunately, the relations adopted by open-source RST parsers are not unified. To allow for comparison against other parsers, 
we do not differentiate subtle differences between relations, therefore grouping very similar relations, following the approach in \cite{feng-hirst-2012-text}. 
(E.g., we consider both Topic-Shift and Topic-Drift to be a \texttt{Topic-Change}). Specifically, this approach does not differentiate between the sequence of nucleus and satellite (e.g., NS-Evaluation and SN-Evaluation are both considered as an \texttt{Evaluation}).

\begin{figure*}[t]
    \centering
    \includegraphics[width=\linewidth]{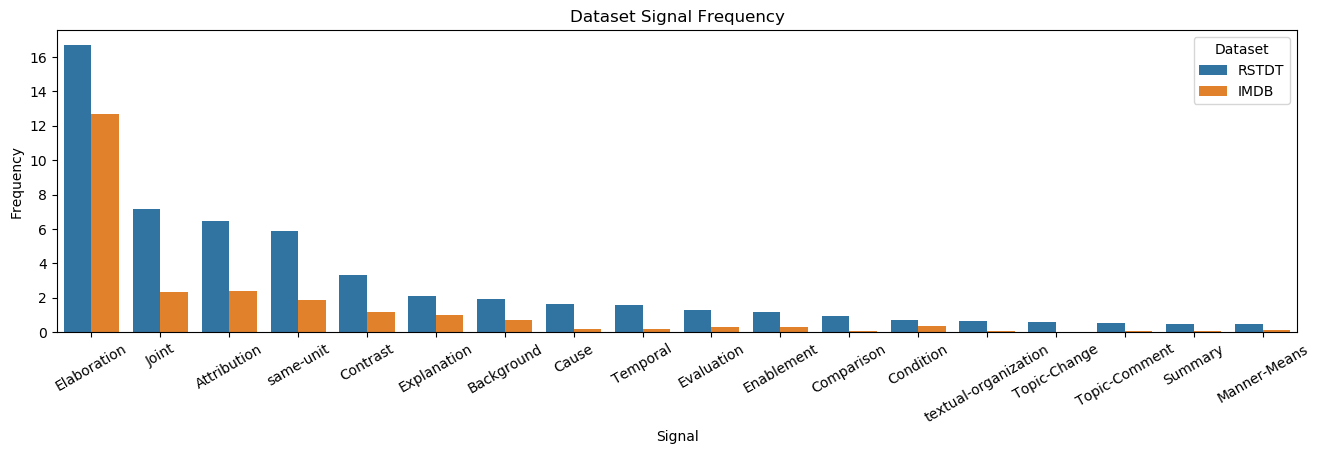}
    \caption{RST relation occurrences per document. RST-DT contain longer documents than IMDB on average. However, the distributions of frequencies between these two datasets are relatively consistent, with \texttt{Elaboration}, \texttt{Joint}, and \texttt{Attribution} the most frequent signals.}
    \label{fig:rst_mean}
\end{figure*}

\paragraph{Tree property features (\texttt{Tree})} We compute the depth and the Yngve depth (the number of right-branching in the tree) \citep{yngve1960model} of each tree node, and include their mean and variance as characteristic features, following previous work extracting tree linguistic features \citep{li-etal-2019-detecting,ConsensusNetworks}.

\paragraph{EDU related features (\texttt{EDU})} We include the mean and variance of EDU lengths of each document. We hypothesize the longer EDUs indicate higher levels of redundancy in discourse, hence extracting rhetorical features require memory across longer spans. 

Overall, there are 24 features from three categories. We normalize them to zero mean and unit variance, and take these RST features for probing. The features are not independent of each other. Specifically, the features of each group tend to describe the same property from different aspects.\footnote{For example, \texttt{Sig} features describe the composition of the document in a histogram. For the same document, if a relation is changed, e.g., from \texttt{Contrast} to \texttt{Attribution}, then the occurrence of both \texttt{Contrast} and \texttt{Attribution} are affected.}

\subsection{Probe}
Our probing method contains two weight parameters, $W_d$ and $W_p$. First, we embed a document with $L$ tokens using a neural LM with $D$ dimensions to get a raw representation matrix $X\in \mathbb{R}^{L\times D}$. We use a projection matrix $W_d \in \mathbb{R}^{D\times d}$ to reduce the embedding dimension from $D$ (e.g., $D=768$ for BERT and $2048$ for XLM) to a much smaller one, $d$. Then, we use self attention similar to \citet{lin2017selfAttention} to collect the information spread across the document to a condensed form:
$$A = (XW_d)^T (XW_d) \in \mathbb{R}^{d\times d}$$

We flatten $A$ into a vector with fixed size: $\tilde{A} = (d^2, 1)$. We use a probing matrix $W_p \in \mathbb{R}^{d^2\times m}$ to extract RST features $v \in \mathbb{R}^{m}$ from attention, normalize them to zero mean and unit variance, and optimize based on the expected $L_2$ error:
$$\min_{W_d,W_p} \mathbb{E} || W_p^T\tilde{A} - v ||^2$$

Note that the reduction from $D$ to $d$ using $W_p$ is necessary, because it significantly lowers the number of parameters of the probing model. If there were no $W_d$ (i.e., $d=768$), then $W_p$ alone would require $768^2 m$ parameters to probe $m$ features. Now, we let $d=10$, then $W_d$ and $W_p$ combined have $D\times d+d^2m \approx 7680+100m$ parameters. Considering $m\in \mathcal{O}(10^1)$, the total parameter size is reduced from $\mathcal{O}(10^6)$ to $\mathcal{O}(10^3)$.  

There is one more step before we can use this loss to measure the difficulty of probing rhetorical features. $L_2$ error scales linearly with the dimension of features $m$, so it is necessary to normalize the $L_2$ error by $m$, to ensure that the losses can be compared across linguistic feature sets. The \emph{difficulty} of probing a group of $m$ features $v\in \mathbb{R}^{m}$ therefore is:
$$\text{Difficulty}=\frac{1}{m} \mathbb{E}\left|\left|W_p^{T} \tilde{A} - v\right|\right|^2$$

\begin{figure*}[t]
    \centering
    \includegraphics[width=\linewidth]{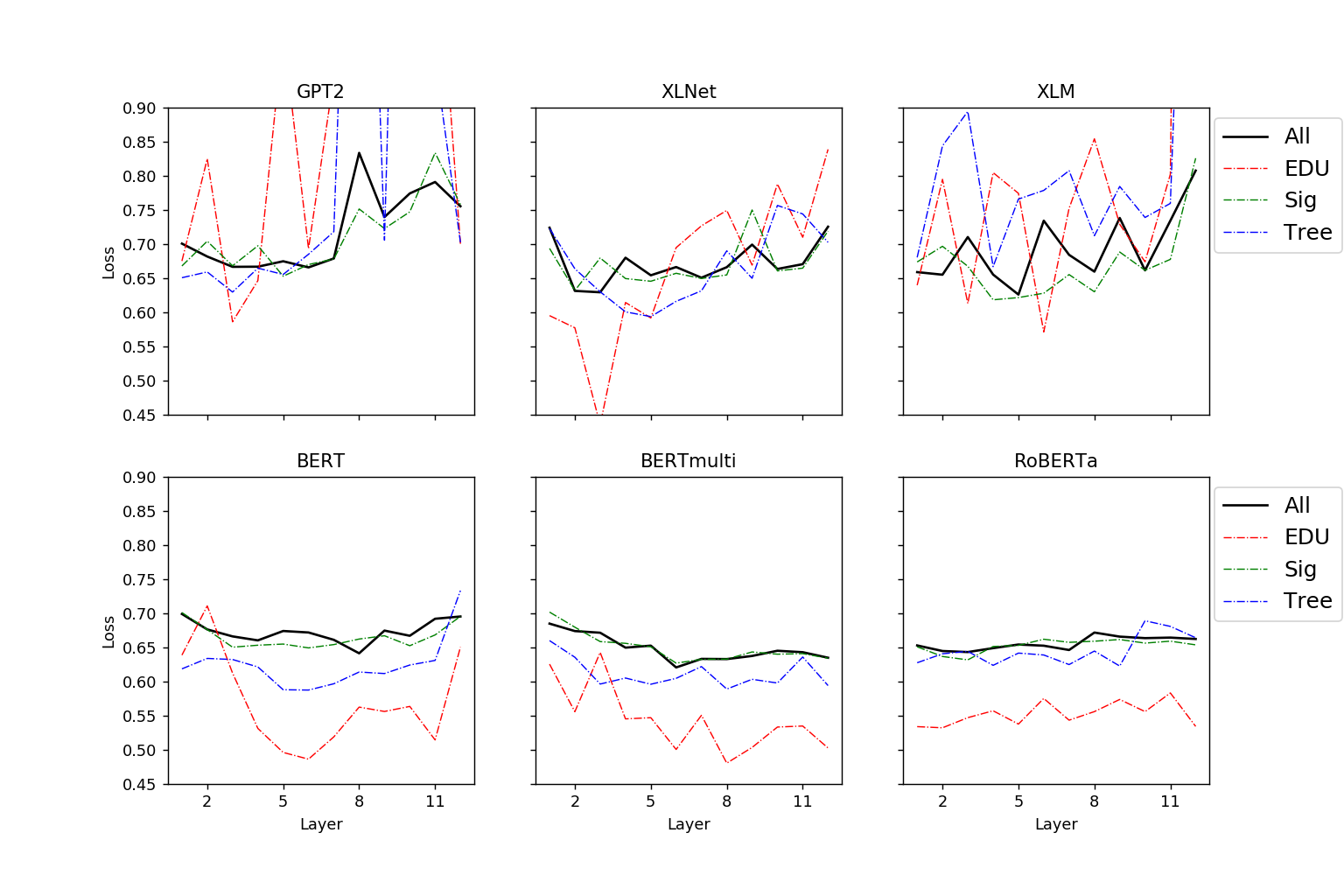}
    \caption{Loss vs layer plot of six neural LMs on four RST feature sets on IMDB. The solid lines represent all RST features combined, while each dash-dotted line denotes one component (EDU, Sig, or Tree feature group for red, green, and blue respectively). In general, BERT-based LMs (BERT, BERT-multi, RoBERTa) encode rhetorical features in a more stable and easy-to-probe manner than the rest.}
    \label{fig:layerplot}
\end{figure*}
\begin{figure*}[t]
    \centering
    \includegraphics[width=\linewidth]{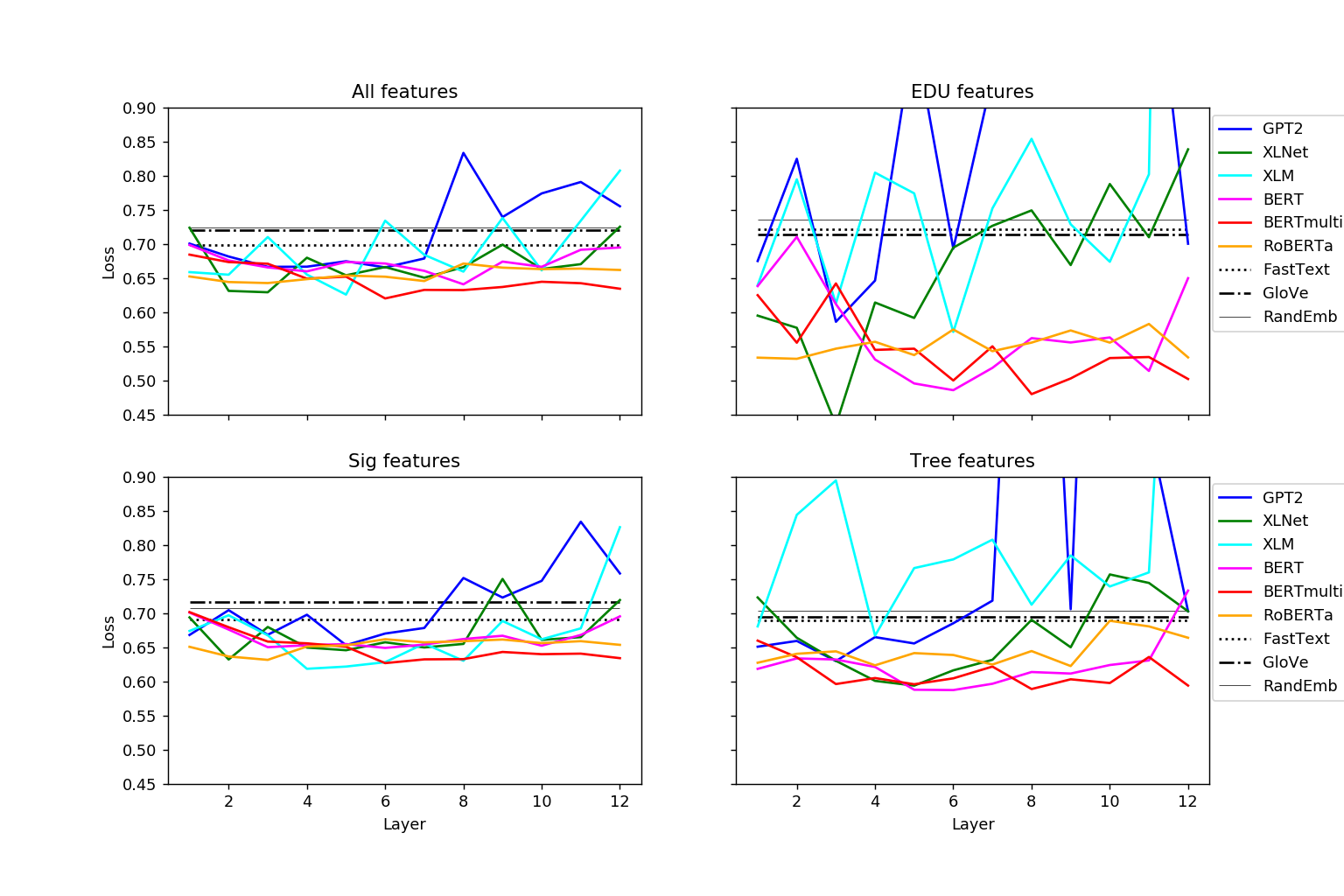}
    \caption{Probing loss, compared to those from non-contextualized baselines, for four feature groups, on IMDB. BERT-based neural LMs stably outperform the word embedding baselines in almost all layers.}
    \label{fig:titleplot}
\end{figure*}

\section{Experiments}\label{sec:Experiments}

\subsection{Data}
Most state-of-the-art rhetorical parsers are trained on either Penn Discourse Treebank \citep{ji-eisenstein-2014-representation,feng-hirst-2012-text} or RST-DT \citep{feng-hirst-2014-linear,joty-etal-2015-codra,Surdeanu2015TwoRSTParsers,Heilman2015FastRS,li-etal-2016-discourse,Wang2017,yu-etal-2018-transition}. Although the documents contain accurate discourse annotations, RST-DT \citep{Carlson2001} only has 385 documents. The Penn Discourse Treebank \citep{prasad2008PDTB} has 2,159 documents but their annotations do not follow the RST framework. So in addition to RST-DT, we extend the analysis to a 100 times larger dataset, IMDB \citep{Maas2011IMDB}.

IMDB contains 50,000 movie reviews without discourse annotations. In these reviews, the authors explain and elaborate upon their opinions towards certain movies and give ratings. 
We removed html tags, and attempt to parse all of them (i.e., both train and test data) using a two-pass parser from \citet{feng-hirst-2014-linear}.
We discarded 1,977 documents that the RST parser generate ill-formatted trees\footnote{As determined by \texttt{nltk.tree}.}. Of the remaining documents, we additionally filtered out those with sequence lengths greater than 512 tokens\footnote{As determined by any one of the tokenizers, since these language models come with their own tokenizers. Note that RoBERTa adds two special tokens, so this threshold becomes 510 for RoBERTa.}, resulting in 40,833 documents.

After parsing each document into an ``RST-tree'', we extracted the features mentioned in Section \ref{subsec:rhetorical-features} from these parsed trees. Figure \ref{fig:rst_mean} shows the occurrence of the 18 RST relations per document, and Table \ref{tab:features_stats} shows the statistics of remaining 6 features. In addition, we include several examples of parsed RST trees in Appendix.

\begin{table}[h]
    \centering
    \begin{tabular}{|l c|}
        \hline
        Feature name & Mean $\pm$ stdev \\ \hline 
        tree\_depth\_mean & 3.9$\pm$1.4 \\
        tree\_depth\_var & 4.6$\pm$4.2 \\
        tree\_Yngve\_mean & 9.2$\pm$8.8 \\
        tree\_Yngve\_var & 100.6$\pm$164.6 \\
        edu\_len\_mean & 8.6$\pm$1.4 \\
        edu\_len\_var & 21.8$\pm$16.0 \\ \hline 
    \end{tabular}
    \caption{Statistics of the 6 non-occurrence-based RST features. The prefix ``tree\_'' here refers to the parsed ``RST-tree''.}
    \label{tab:features_stats}
\end{table}

\subsection{Language models} 
We considered the following popular neural LMs:
\begin{itemize}[noitemsep]
    \item $\text{BERT}_{\text{BASE}}$ \citep{devlin2019bert} This LM with 110M parameters is built with 12-layer Transformer encoder \citep{Vaswani2017} with 768 hidden dimensions. It is trained with masked LM (i.e., cloze) and next sentence prediction objectives using 16GB text.
    \item BERT-multi \citep{HuggingfaceTransformers2019} Same as BERT, BERT-multi is also a 12-layer Transformer encoder with 768 hidden dimensions and 110M parameters. Its difference from BERT is that, BERT-multi is trained on top 104 languages with the largest Wikipedia.
    \item RoBERTa \citep{Liu2019roberta} is an enhanced version of BERT with the same architecture, similar masked LM objectives, and 10 times larger training corpus (over 160GB).
    \item GPT-2 \citep{radford2019language} is a 12-layer Transformer decoder with 768 hidden dimensions. There are 117M parameters in total. GPT-2 is pretrained on 40GB of text. Unlike BERT, GPT-2 is a uni-directional LM. 
    \item XLM \citep{Lample2019} is 12-layer Transformer with 2048-hidden dimensions. We use the English model trained with masked language model (MLM) objective. Different from BERT (taking sentence pairs as input), XLM takes continuous streams of tokens as input.
    \item XLNet \citep{Yang2019} is a 12-layer Transformer-XL \citep{dai2019transformerXL} with two streams of self attention and 768 hidden dimensions and 110M parameters. The XLNet we use is trained on 33GB texts using the ``permutation language modeling'' objective, with its LM factorization according to shuffled orders, but its positional encoding correspond to the original sequence order. The permutation LM objective introduces diversity and randomness to the context.
\end{itemize}
To make comparisons between models fair, we limit to 12-layer neural LMs. The models are pretrained by Huggingface \citep{HuggingfaceTransformers2019}.

\subsection{Implementation}
We formulated probing as an optimization problem, and implemented our solution with PyTorch \citep{paszke2019pytorch} and the Adam optimizer \citep{kingma2014adam} for 40 epochs. If the training loss stalls (i.e., does not change by $\geq 10^{-3}$), or if the training loss rises by more than 10\% from the previous epoch, we stop the optimization. All optimizations follow the same learning rate tuning schemas.

In our experiments, the representation dimension $d$ is taken to be 10, while the LM dimensions $D$ is $2048$ for XLM and $768$ for the rest.

\section{Results and Discussion}

\begin{figure*}
    \centering
    \includegraphics[width=\linewidth]{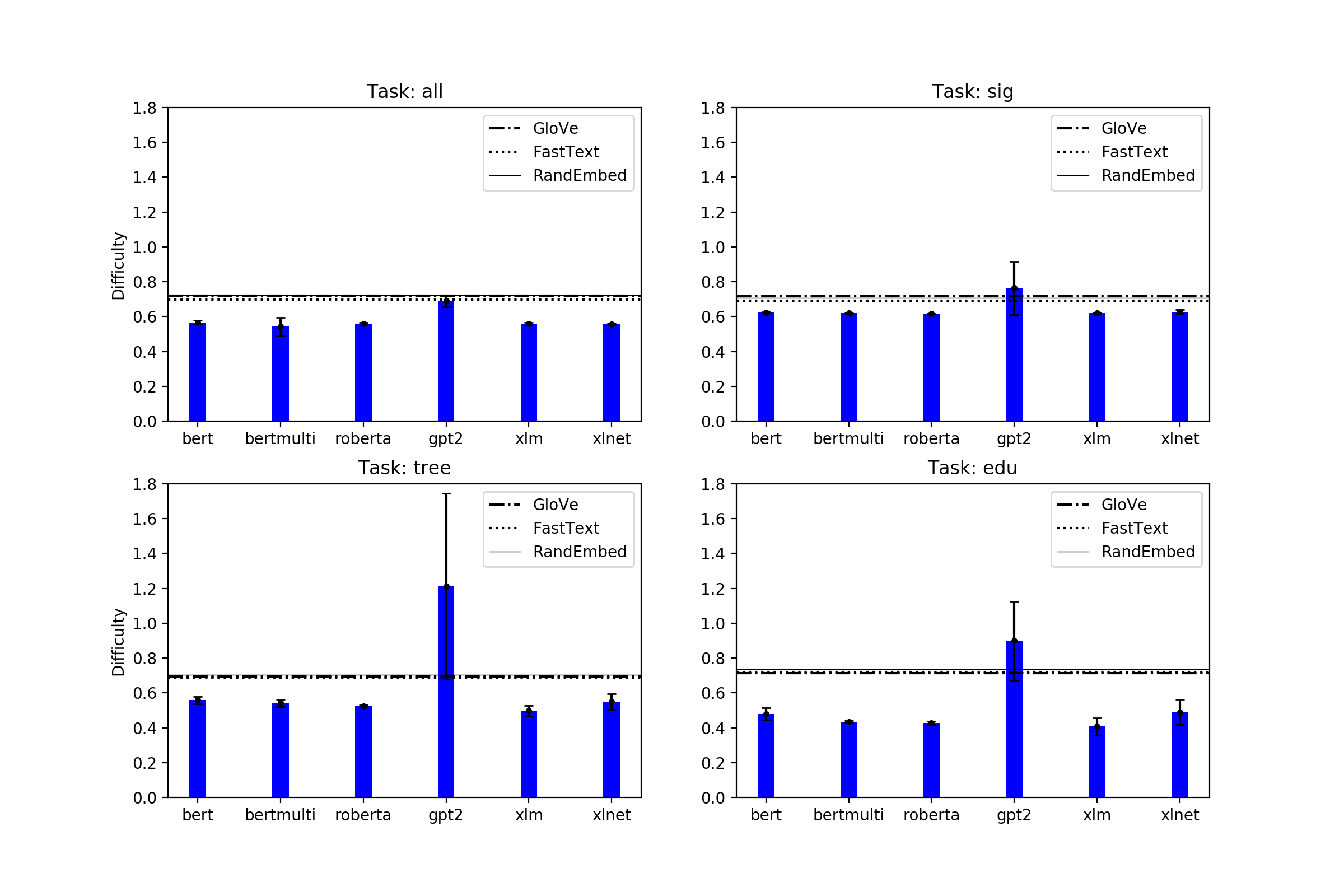}
    \caption{Probing performances of averaging 12 layers for 6 neural LMs on 4 tasks in IMDB, compared to the three non-contextual baselines. All LMs except GPT-2 outperform non-contextual LM baselines. Plots for RST-DT (Figure \ref{fig:avg_plot_rstdt} in Appendix) reveal similar patterns.}
    \label{fig:avg_plot}
\end{figure*}

\subsection{Where do LMs encode RST features?}
From Figure \ref{fig:layerplot}, neural LMs encode RST features in different manners, depending on their structures. In general, for BERT-based models, features seem to distribute evenly across layers. On GPT-2 and XLNet,
lower layers seem to encode slightly more \texttt{EDU} and \texttt{Sig} features than higher levels, whereas Tree features seem to be more concentrated in layers 2-6.
The results on XLM are relatively noisy, possibly because the uni-language version does not benefit from the performance boost of cross-language modeling.

Contrasting with previous work that suggested that middle layers most contain syntactic features \citep{hewitt-manning-2019-structural-probe,Jawahar2019}, our results indicate a less definitive localization for discourse features, except for the first and final layers. We suggest that the reason they encode less discourse information is that the first layer focuses on connections between ``locations'', while the final layer focuses on extracting representations most relevant to the final task.

\paragraph{Are RST features equally hard to probe?}
Figure \ref{fig:layerplot} also shows the difficulty in probing features across feature sets. In BERT-based models, \texttt{EDU} and \texttt{Tree} features are comparably easier to probe, whereas the \texttt{Sig} feature groups is more challenging.  
However, GPT-2, XLNet, and XLM do not regard \texttt{EDU} or \texttt{Tree} features easier to probe than other groups.
Nevertheless, the results on all features correlate more to the \texttt{Sig} features.

\paragraph{How about averaging layers?}
For comparison, we also used the mean of all 12 layers for each neural LM. Figure \ref{fig:avg_plot} shows the probing results. Except GPT-2, other LMs show similar performances when the representations of layers are averaged. In addition, the performances show that \texttt{Sig} features are harder to probe than \texttt{Tree} and \texttt{EDU} features, whereas the aggregation task (using all features) appears harder than each of its three component feature groups.

\subsection{Deconstructing the probe}
We perform ablation studies to illustrate the effectiveness of probing, deconstructing the language model probe step-by-step. First, we get rid of the contextualization component in language modeling by using non-contextualized word embeddings, GloVe and FastText. Then, we discard the semantic component of word embedding by mapping tokens to randomly generated vectors (RandEmbed). Finally, we remove all information pertaining to the text, leading to a random predictor for RST features, RandGuess.

\paragraph{Non-contextualized word embeddings} We consider two popular word embeddings here:
\begin{itemize}[noitemsep]
    \item GloVe \citep{pennington2014glove} contains 2.2M vocabulary items and produces 300-dimensional word vectors. The GloVe embedding we use is pretrained on Common Crawl.
    \item FastText \citep{bojanowski2017enriching} is trained on Wikipedia 2017 + UMBC (16B tokens) including subword information, and produces 300-dimensional word vectors.
\end{itemize}
Word embeddings map each token into a $D$-dimensional semantic space. Therefore, for a document of length $L$, the embedded matrix also has shape $L\times D$. The difference from the contextualized neural LMs is that, the $D$-dimensional vectors of every word do not depend on their contexts.

\paragraph{Random embeddings}
In this step, we assign a non-trainable random embedding vector per token in the vocabulary. This removes the semantic information encoded by GloVe and FastText word embeddings.

As shown in Figure \ref{fig:titleplot}, \ref{fig:avg_plot}: RandEmbed is worse than GloVe and FastText (except for GloVe in \texttt{Sig} features task). This verifies some semantic information is preserved in word embeddings. 

\paragraph{Contextualized LMs against baseline}
First, the lack of context restrict the probing performance of non-contextualized baselines. They are worse than most layers in contextualized LMs (in Figure \ref{fig:titleplot}), and are worse than all except GPT-2 if we average the layers (in Figure \ref{fig:avg_plot}).

Second, it is impossible for any LM to have a ``negative'' rhetorical capacity. If the probing loss is worse than RandEmbed baseline, that means the RST probe can not detect rhetorical features of the given category encoded in the representations.
This is what happens in some layers of GPT-2, XLM, and XLNet, and the mean of all layers of GPT-2.

\paragraph{Random guesser} 
To measure the capacity of baseline embeddings, we set up a random guesser as a ``baseline-of-baseline''.
The random guesser outputs the arithmetic mean of RST features plus a small Gaussian noise (with s.d. $\sigma\in \{0, 0.01, 0.1, 1.0\}$)  The output of RandGuess is completely independent of the discourse. As shown in Table \ref{tab:random-guessor}, the best of the four random guessers is much worse than any of the three word embedding baselines, which is expected. 

\begin{table}[h]
    \centering
    \begin{tabular}{|l| c c c c|}
        \hline 
        \multirow{2}{*}{Config} &  \multicolumn{4}{c|}{RST Feature Set} \\ \cline{2-5}
        & All & EDU & Sig & Tree \\ 
        \hline 
        FastText & .6987 & .7215 & .6911 & .6889 \\
        GloVe & .7204 & .7142 & .7166 & .6942 \\
        RandEmbed & .7238 & .7365 & .7077 & .7034 \\ \hline
        RandGuess & 1101.5 & 128.9 & 3.1 & 6799.0 \\ \hline 
    \end{tabular}
    \caption{Comparison between RST probing losses of non-contextual word embeddings (FastText, GloVe), random embedding (RandEmbed), and a trivial guessor (RandGuess).}
    \label{tab:random-guessor}
\end{table}

\subsection{Why are some LMs better?}
From probing experiments (Figure \ref{fig:layerplot}, \ref{fig:titleplot}, and \ref{fig:avg_plot}) we can see that BERT-based LMs have slightly better rhetorical capacities than XLNet, and much better capacities than GPT-2. We present two hypotheses as following.

\paragraph{Rhetorics favor contexts from both directions} BERT-based LMs use Transformer encoders, whereas GPT-2 use Transformer decoders. Their main difference is that a Transformer encoder considers contexts from both ``past'' and ``future'', while a Transformer decoder only conditions on the context from the ``past'' \citep{Vaswani2017}. GPT-2 attends to uni-directional contexts. Apparently both the ``past'' and ``future'' context would contribute to the rhetorical features of words. Without ``future'' contexts, GPT-2 would encode less rhetorical information.

\paragraph{Random permutation makes encoded rhetorics harder to decode} The difference between XLNet and other LMs is the permutation in context. While permutation increases the diversity in discourse, they could also bring in new meaning to the texts. For example, the sentence in Figure \ref{fig:rst-tree} (``I didn't know this is from C, but it is very good!'') has several syntactically plausable factorization sequences:
\begin{itemize}[nosep]
    \item I didn't know C ...
    \item ... this is C ...
    \item I know it is very good ...
    \item I didn't know this is good ...
    \item ... didn't this C good ...
\end{itemize}
Apparently such diversity in contexts makes the upper layers of XLNet contain harder-to-decode rhetorical features. If we average the representations of all layers, XLNet has larger variance than BERT-based LMs. We hypothesize that larger layer-wise difference is a factor of such instability for averaged representations.

\subsection{Limitations}
RST probing is not perfect. While we designed our comparisons to be rigorous, there are still several limitations to the RST probe, described below.
\begin{itemize}[nosep]
\item{RST signals are noisy.} The RST relation classification task is less defined than established tasks like POS tagging. Humans tend to disagree with the annotators, resulting in a merely 65.8\% accuracy in relation classification (i.e., the task introduced by \citet{Marcu2000TheoryPracticeDiscourse}). Regardless, state-of-the-art discourse parsers currently have performances slightly higher than 60\% \citep{feng-hirst-2014-linear,ji-eisenstein-2014-representation,Wang2017}.

\item{Train / test corpus discrepancy of RST parsers.} Most available RST parsers are trained on RST-DT consisting of Wall Street Journal articles. The results of parsers are affected by the corpus. As shown in some examples in Appendix, the IMDB movie review dataset contains less formal languages, introducing noise in segmentations and relation signals. To counteract noise of this type, we recommend evaluating LMs using a corpus similar to the scenario of applying the LM.

\item Only 12-layer LMs are involved, to compare across various layers fairly. But our approach would be applicable to 3-layer ELMo and deeper LMs as well. Appropriate statistical controls would naturally need to be applied.

\item{Not all documents can be analyzed.} First, documents longer than 512 tokens cannot be encoded into one vector in our probing model. Second, while RST provides elegant frameworks for analyzing rhetorical structures of discourse, in practice, the RST pipeline does not guarantee a successful analysis for an arbitrary document scraped online.
\end{itemize}

\section{Related work}
Recent work has considered the interpretability of contextualized representations. For example, \citet{Jain2019AttentionIN} found attention to be uncorrelated to gradient-based feature importance, while \citet{wiegreffe-2019-attention} suggested such approaches allowed too much flexibility to give convincing results. Similarly, \citet{Serrano2019} considered attention representations to be noisy indicators of feature importance.

Many tasks in argument mining, similar to our task of examining neural LMs, require understanding the rhetorical aspects of discourse \citep{Lawrence2019ArgumentMiningSurvey}. This allows RST to be applied in relevant work.
For example, RST enables understanding and analyzing argument structures of monologues \citep{peldszus-stede-2016-rhetorical} and, when used with other discourse features, RST can improve role-labelling in online arguments \citep{chakrabarty-etal-2019-ampersand}.

Probing neural LMs is an emergent diagnostic task on those models. 
Previous work probed morphological \citep{bisazza-tump-2018-lazy}, agreement \citep{giulianelli-etal-2018-hood}, and syntactic features \citep{hewitt-manning-2019-structural-probe}. 
\citet{hewitt-liang-2019-designing} compared different probes, and recommended linear probes with as few parameters as possible, for the purpose of reducing overfitting. Recently, \citet{pimentel2020information} argued against this choice from an information-theoretic point of view. \citet{voita2020information} presents an optimization goal for probes based on minimum description length.

\citet{Liu2019} proposed 16 diverse probing tasks on top of contextualized LMs including token labeling (e.g., PoS), segmentation (e.g., NER, grammatical error detection) and pairwise relations. While LMs augmented with a probing layer could reach state-of-the-art performance on many tasks, they found that LMs still lacked fine-grained linguistic knowledge.
DiscoEval \citep{Chen2019} showed that BERT outperformed traditional pretrained sentence encoders in encoding discourse coherence features, which our results echo.

\section{Conclusion}
In this paper, we propose a method to quantitatively analyze the amount of rhetorical information encoded in neural language models. We compute features based on Rhetorical Structure Theory (RST) and probe the RST features from contextualized representations of neural LMs. Among six popular neural LMs, we find that contextualization helps to generally improve the rhetorical capacities of LMs, while individual models may vary in quality. In general, LMs attending to contexts from both directions (BERT-based) encode rhetorical knowledge in a more stable manner than those using uni-directional contexts (GPT-2) or permuted contexts (XLNet).

Our method presents an avenue towards quantitatively describing rhetorical capacities of neural language models based on unlabeled, target-domain corpus. This method may be used for selecting suitable LMs in tasks including rhetorical acts classifications, discourse modeling, and response generation.

\section*{Acknowledgement}
We thank the anonymous reviewers for feedback. Rudzicz is supported by a
CIFAR Chair in artificial intelligence. Abdalla is supported by a Vanier scholarship.

\bibliographystyle{acl_natbib}
\bibliography{emnlp2020}

\newpage 
\appendix 

\section{Experiments on RST-DT}

\begin{figure*}
    \centering
    \includegraphics[width=\linewidth]{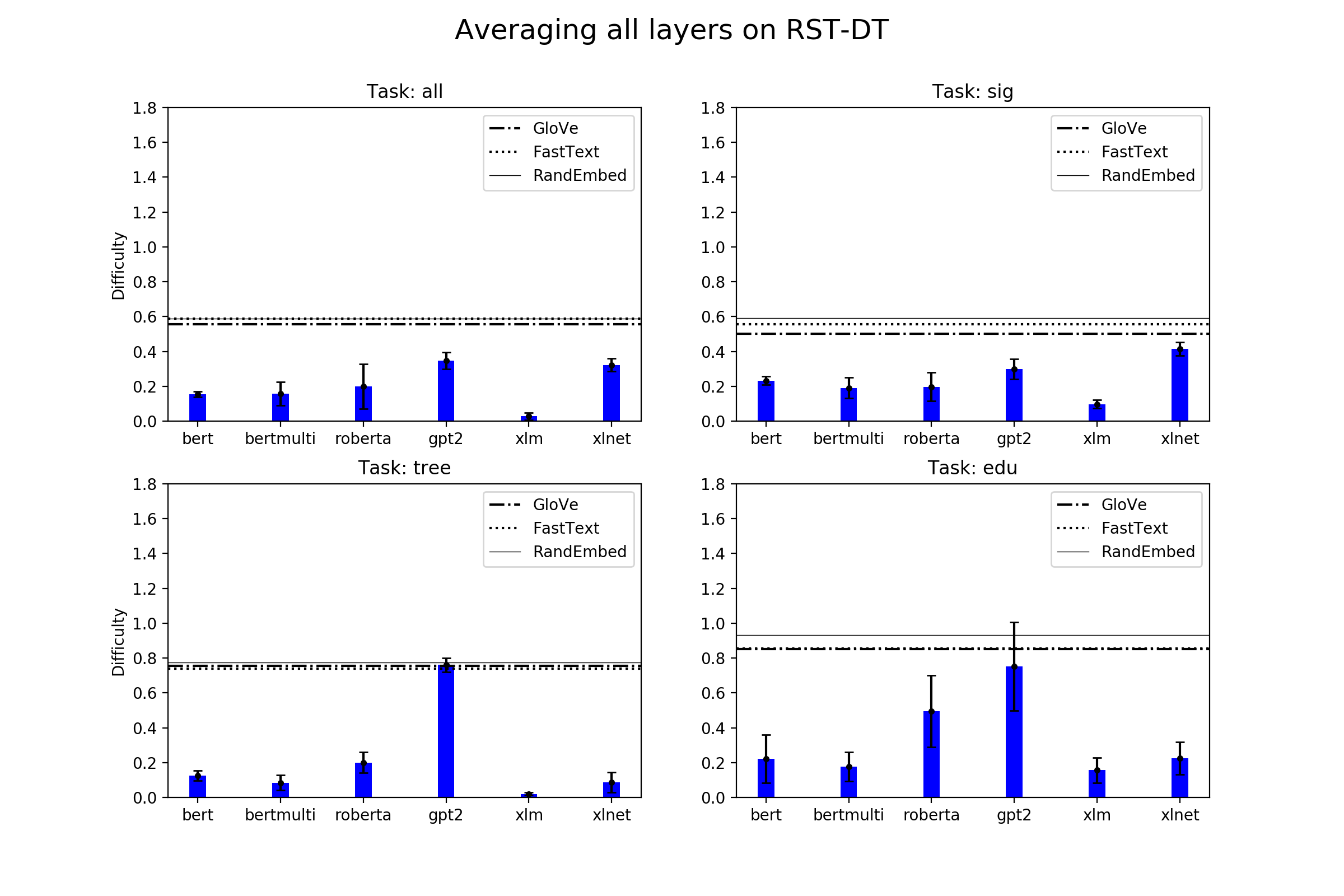}
    \caption{Probing performances of averaging 12 layers for 6 neural LMs on 4 tasks in RST-DT, compared to the three non-contextual baselines.}
    \label{fig:avg_plot_rstdt}
\end{figure*}

As a sanity check, we include experiments on RST-DT \citep{Carlson2001} corpus with the same preprocessing and feature extraction procedures (i.e., perform feature extraction and embedding on the article level, and ignoring the overlength articles). As shown in Figure \ref{fig:avg_plot_rstdt}, BERT-family and XLM outperform GPT-2 and XLNet. Also, the noncontextualized embedding baselines show worse performances than contextualized embeddings in general, with some exceptions (e.g., GPT-2 on EDU features). These are similar to the IMDB results. 

What are different is that the probing losses of RST-DT are lower than the IMDB experiments in general. We consider two possible explanations. First, the IMDB signals contain more noise, so that probing rhetorical features from IMDB would be naturally more difficult than probing from the RST-DT dataset. Second, it is possible that the probes overfit the much smaller RST-DT dataset.

\section{Examples of parse trees}
We include several examples of IMDB parse trees in Appendix here, including some examples where the RST parser makes mistakes on a new domain, movie review. 
For clarity of illustration, these examples are among the shorter movie reviews. More parse trees can be generated by our visualization code, which is contained in our submitted scripts.

\begin{figure*}
    \centering
    \includegraphics[width=\linewidth]{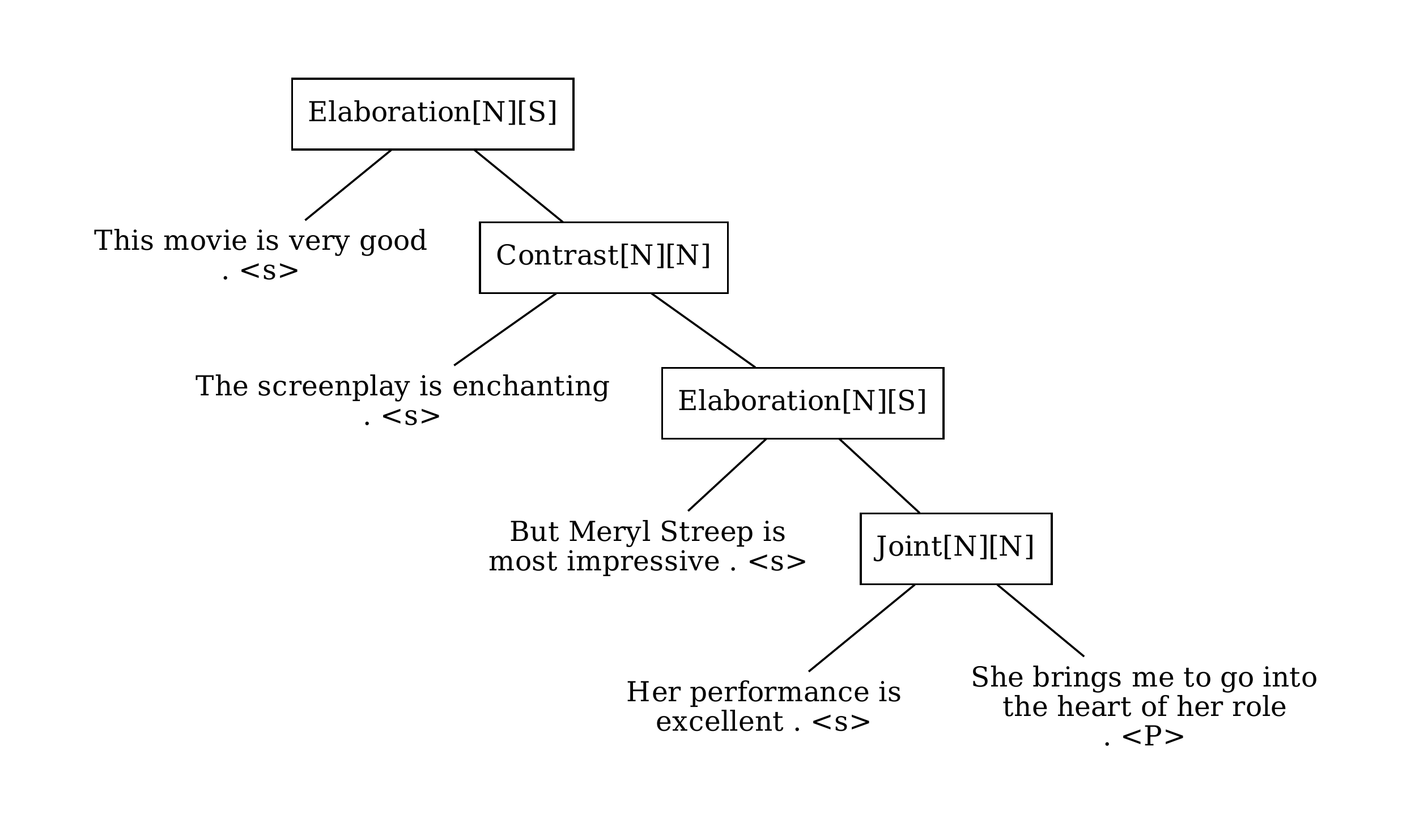}
    \caption{IMDB \texttt{train/pos/10348\_8.txt}. The \texttt{<s>} and \texttt{<P>} are appended automatically by the parser, marking the end of sentences and paragraphs respectively.}
    \label{fig:example_1}
\end{figure*}

\begin{figure*}
    \centering
    \includegraphics[width=\linewidth]{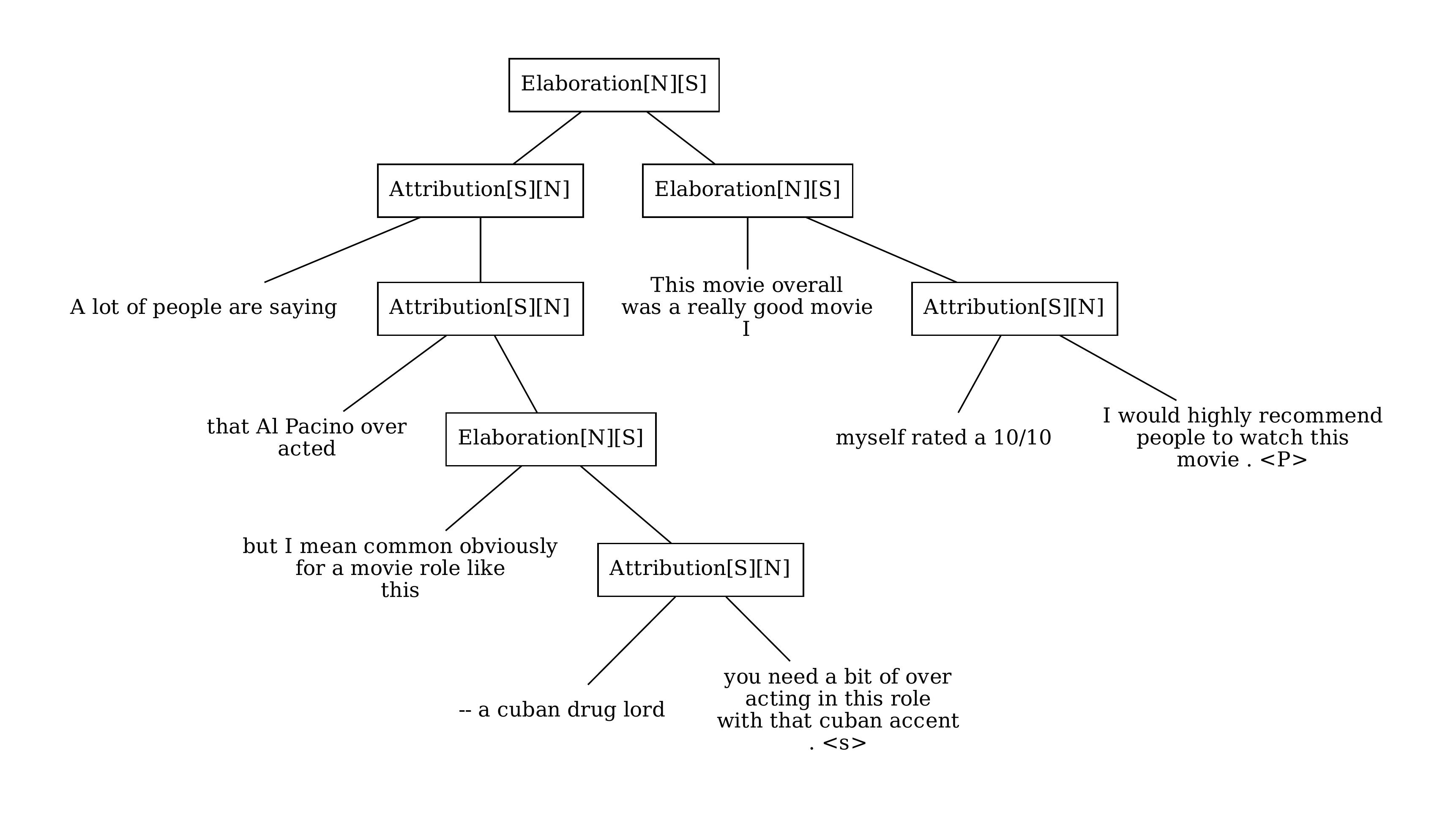}
    \caption{IMDB \texttt{train/pos/11857\_10.txt}. There is an EDU segmentation error: the ``I'' is incorrectly assigned to the previous sentence ``This movie overall was a really good movie''. Apparently some lexical cues the EDU segmentator relies on (e.g., sentence finishes with a period sign) is not always followed in IMDB.}
    \label{fig:example_2}
\end{figure*}

\begin{figure*}
    \centering
    \includegraphics[width=\linewidth]{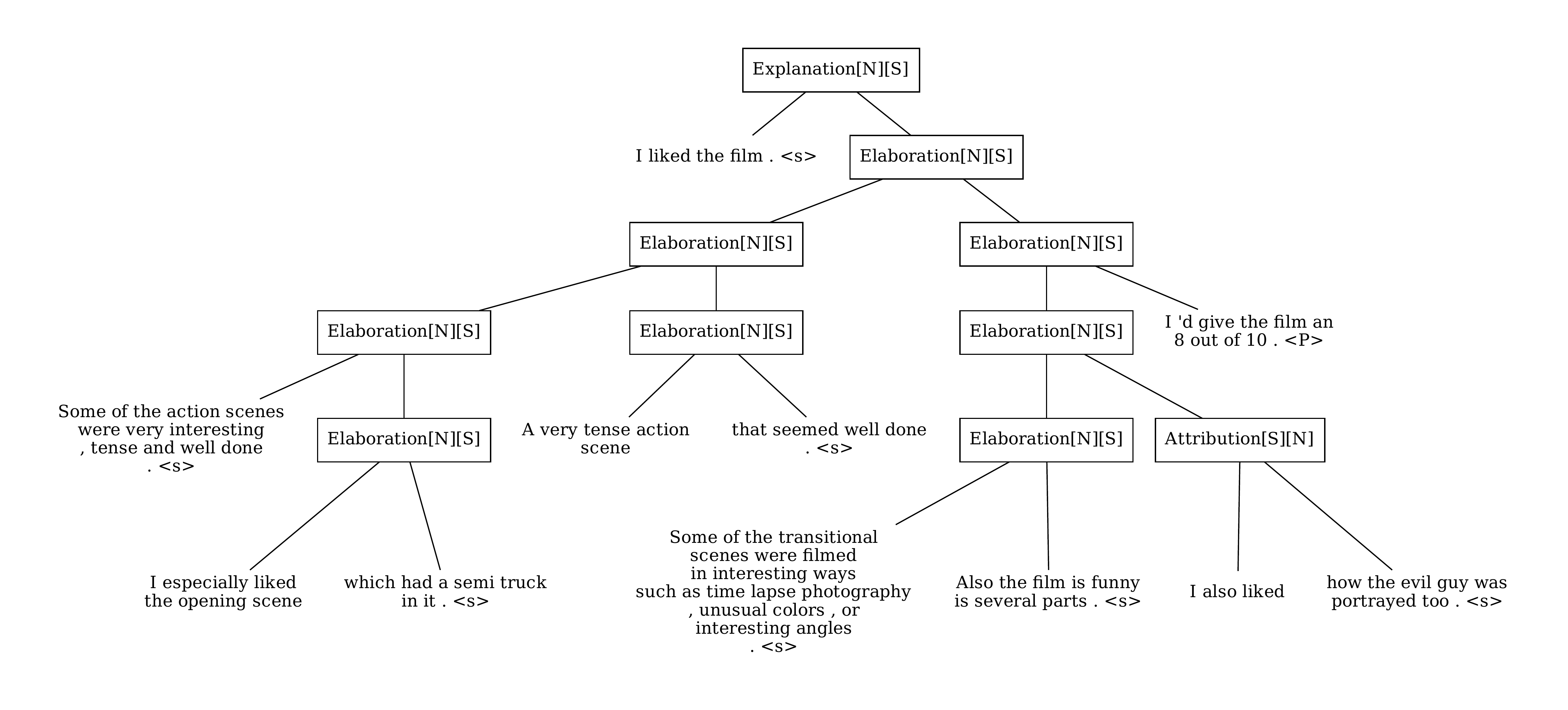}
    \caption{IMDB \texttt{train/pos/1000\_8.txt}. The parser captures the key sentence of this review. All sentences following the first one act as reasons to explain how the reviewer liked the film.}
    \label{fig:example_3}
\end{figure*}

\begin{figure*}
    \centering
    \includegraphics[width=\linewidth]{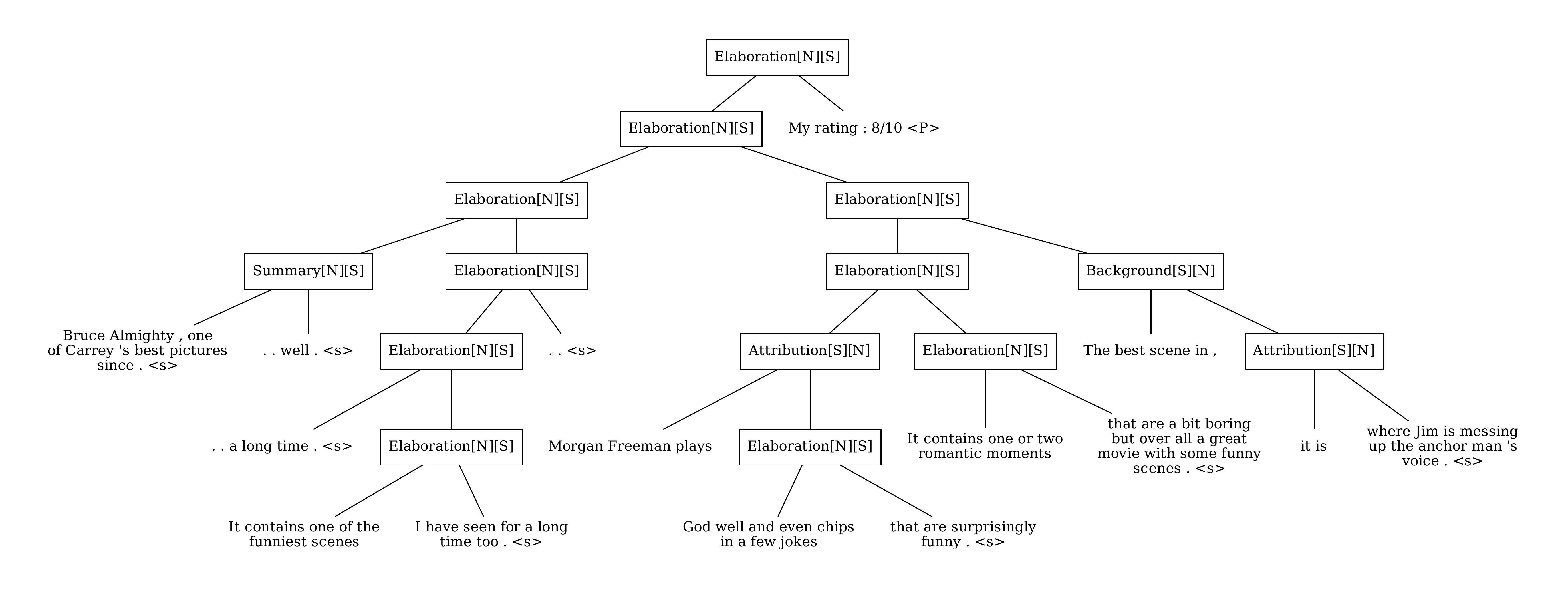}
    \caption{IMDB \texttt{train/pos/10301\_8.txt}. The interjection, ``well'', is incorrectly identified as the satellite of the \texttt{summary} signal. This is likely caused by the discrepancy between the train (RST-DT) and test (IMDB) corpus discrepancy for the RST parser. The RST-DT dataset contains news articles, which are more formal than the online review in IMDB. The term ``well'' is therefore more likely to be identified as other senses.}
    \label{fig:example_4}
\end{figure*}

\begin{figure*}
    \centering
    \includegraphics[width=\linewidth]{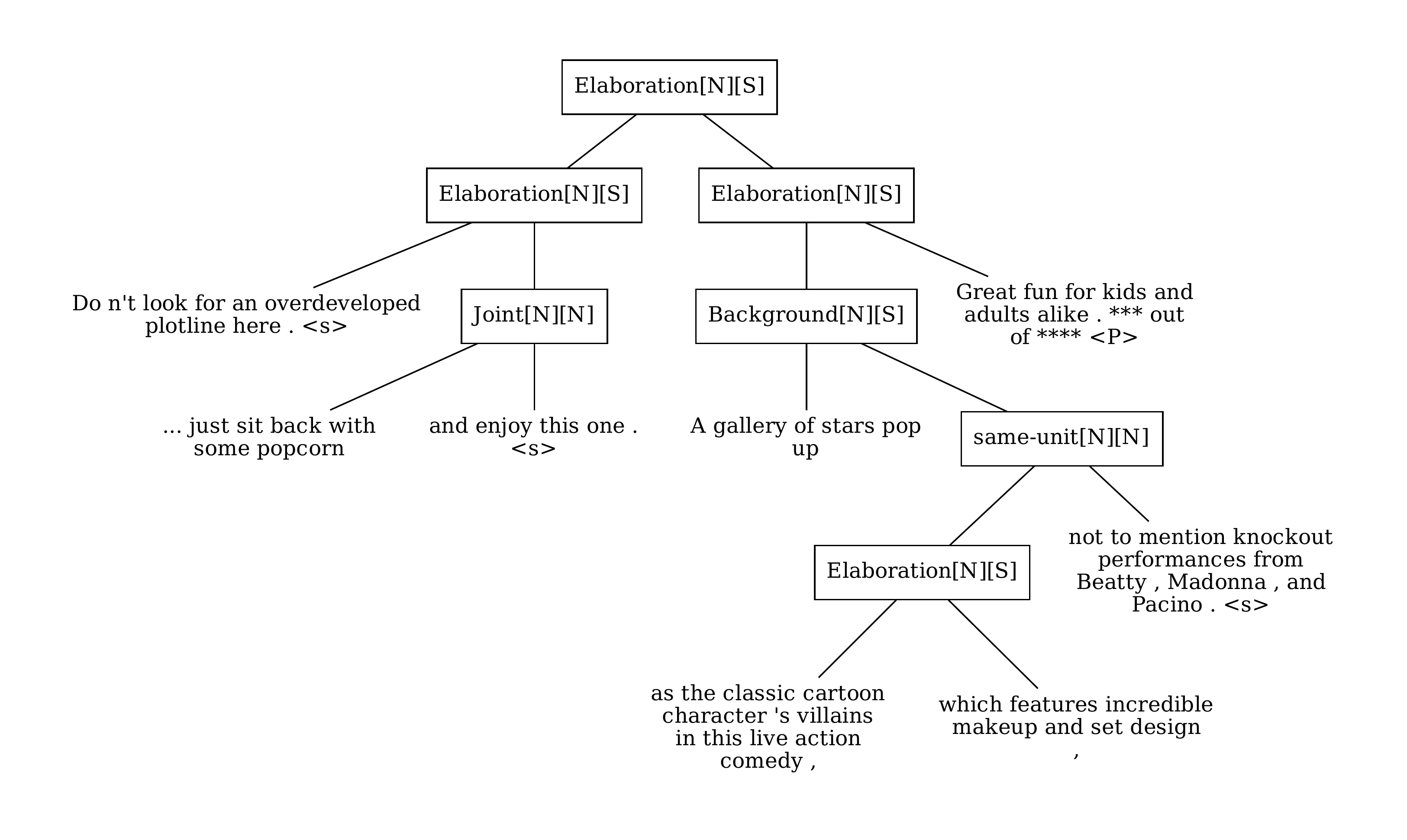}
    \caption{IMDB \texttt{train/pos/11825\_8.txt}. One might suggest that the last EDU could be moved one level higher (so that it summarizes the whole review), but this parsing is also reasonable, since the mention of kids elaborates the descriptions of the makeup and the views.}
    \label{fig:example_5}
\end{figure*}

\begin{figure*}
    \centering
    \includegraphics[width=\linewidth]{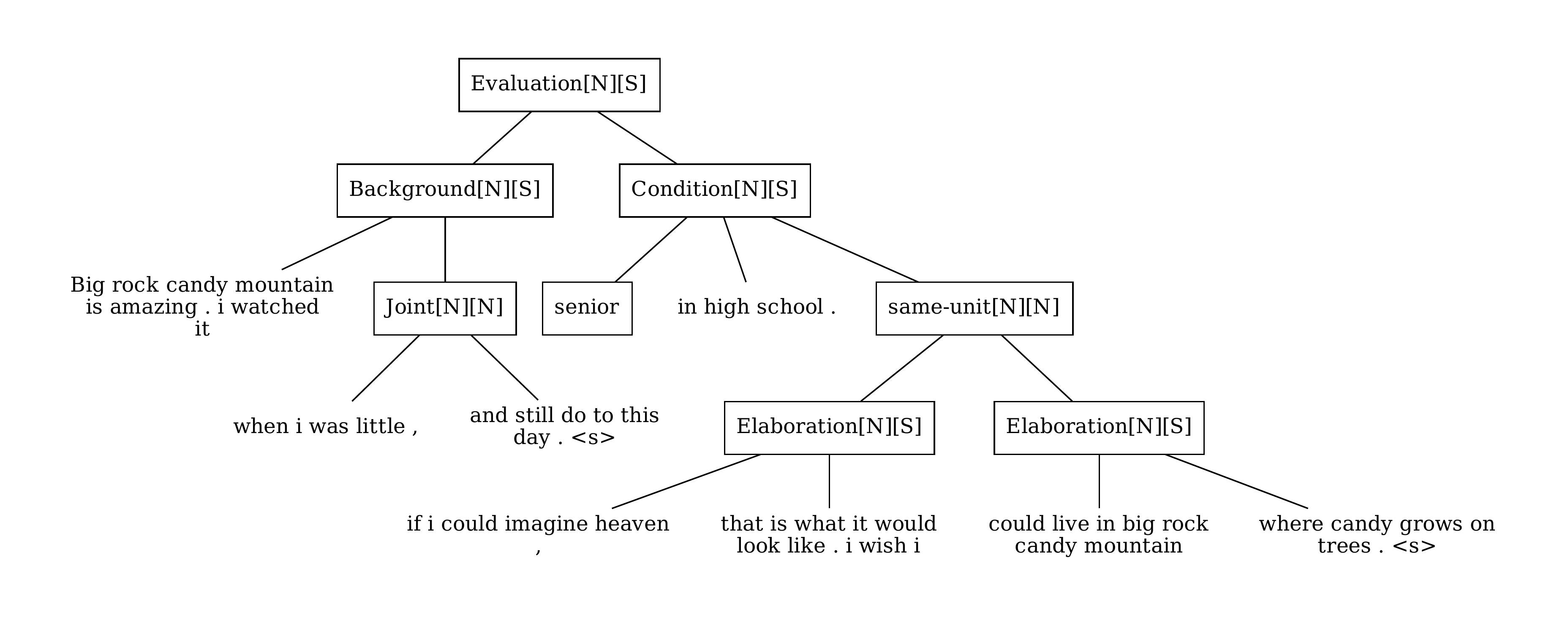}
    \caption{IMDB \texttt{train/pos/10788\_10.txt}. This is an example of the EDU segmentation contains mistake. The ``i wish i'' should be merged with the subsequent EDU, ``could live in big rock candy mountain''. Note that the sentence starts with two lowercase ``i'' (which should be uppercase). The non-standard usages like these are unique for less formal texts like IMDB.}
    \label{fig:example_6}
\end{figure*}

\begin{figure*}
    \centering
    \includegraphics[width=\linewidth]{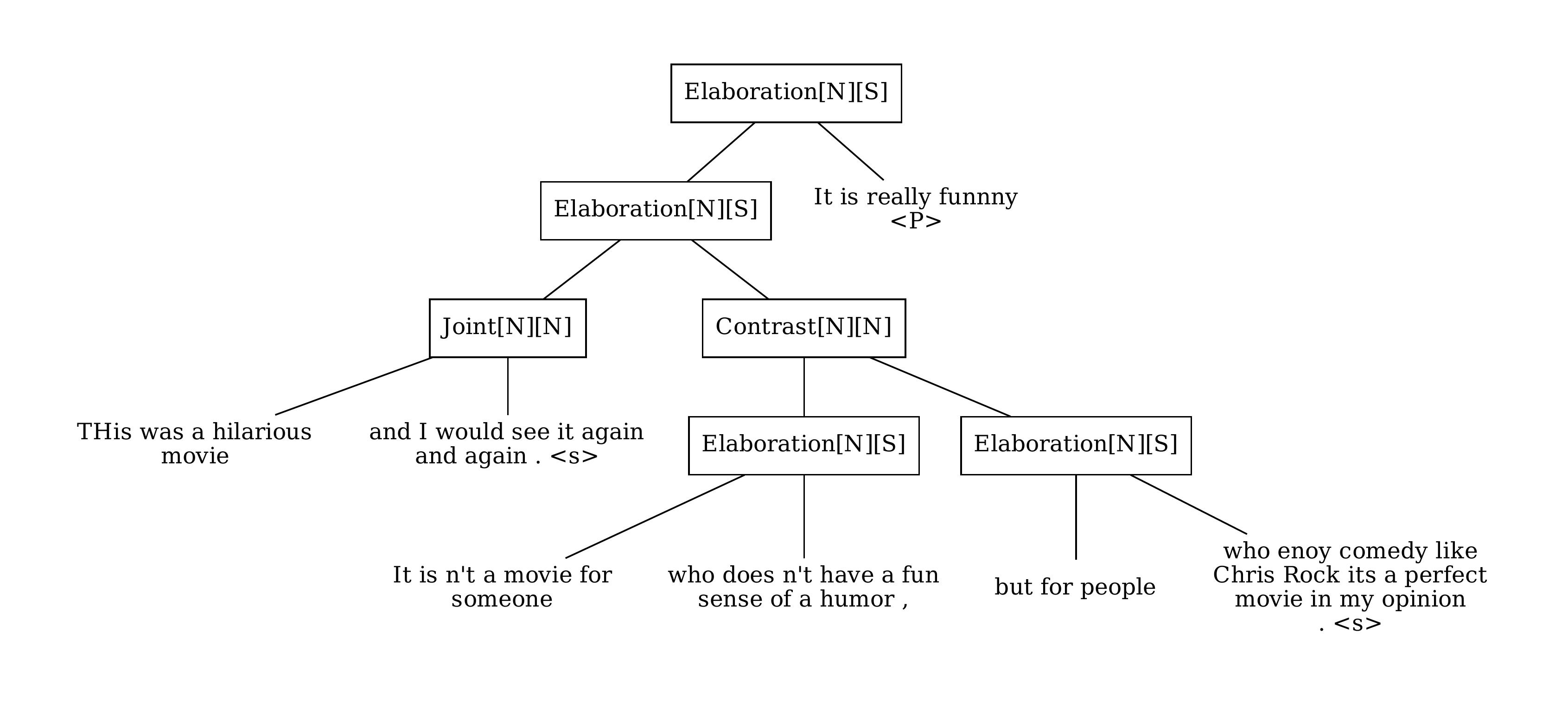}
    \caption{IMDB \texttt{train/pos/11686\_10.txt}.}
    \label{fig:example_7}
\end{figure*}

\begin{figure*}
    \centering
    \includegraphics[width=\linewidth]{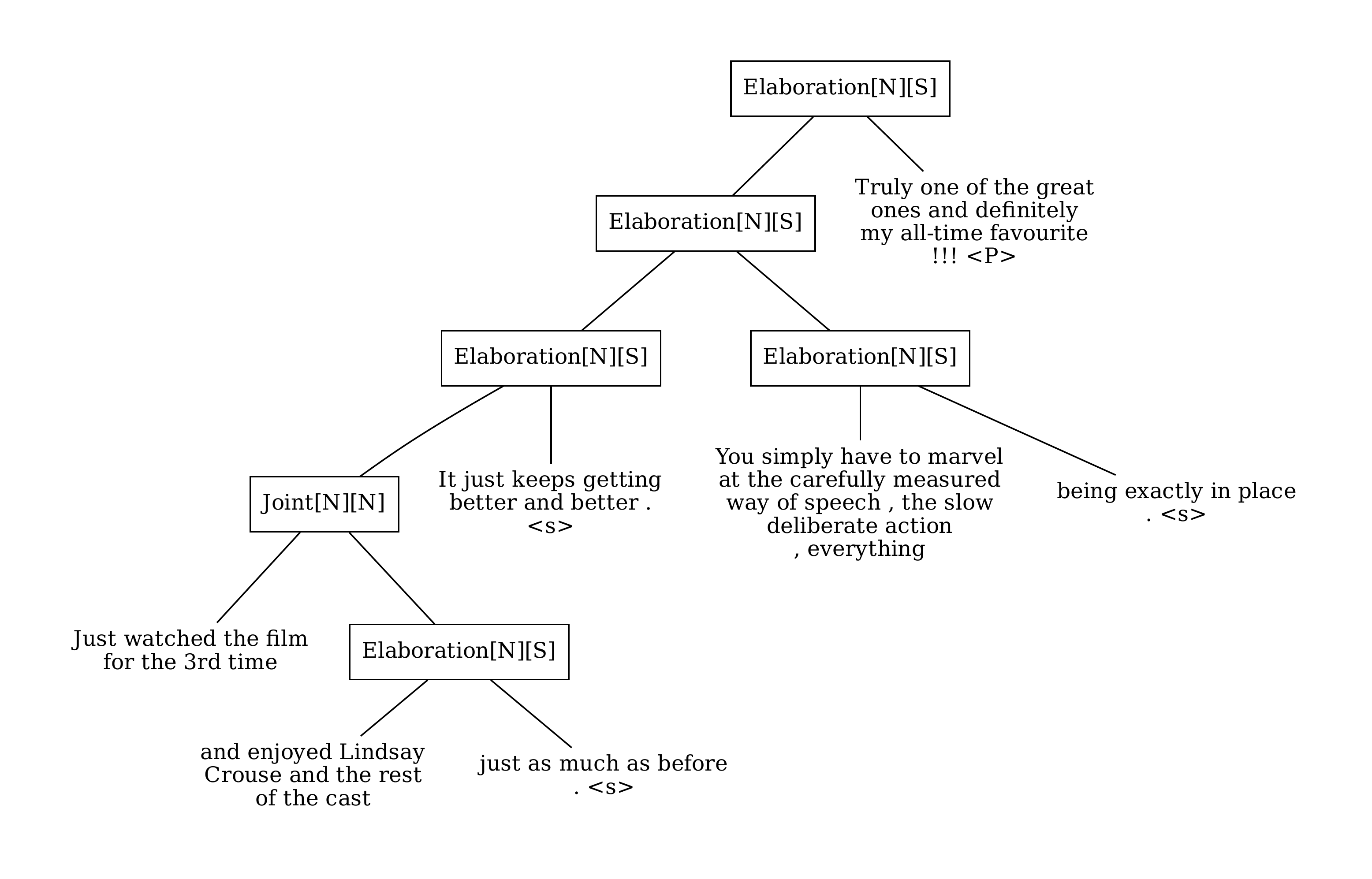}
    \caption{IMDB \texttt{train/pos/10264\_10.txt}}
    \label{fig:example_8}
\end{figure*}

\end{document}